\documentclass[twocolumn,english]{article}
\pdfoutput=1
\usepackage[T1]{fontenc}
\usepackage[latin9]{inputenc}
\usepackage{array}
\usepackage{amsmath}
\usepackage{graphicx}
\usepackage[unicode=true,pdfusetitle,
 bookmarks=true,bookmarksnumbered=false,bookmarksopen=false,
 breaklinks=false,pdfborder={0 0 0},pdfborderstyle={},backref=false,colorlinks=true]
 {hyperref}

\makeatletter

\providecommand{\tabularnewline}{\\}

\usepackage{babel}
\usepackage{caption}
\usepackage{iccv}
\usepackage{times}
\usepackage{epsfig}
\usepackage{cite}
\usepackage{colortbl}

\iccvfinalcopy 

\ificcvfinal\fi

\def\NoSpacing{\itemsep=2pt\topsep=2pt\partopsep=2pt\parskip=2pt\parsep=2pt}

\renewenvironment{thebibliography}[1]{\begin{OrigBiblio}{#1}\NoSpacing}{\end{OrigBiblio}}

\def\myDisplaySkip{0pt}
\abovedisplayskip=\myDisplaySkip
\belowdisplayskip=\myDisplaySkip
\abovedisplayshortskip=\myDisplaySkip
\belowdisplayshortskip=\myDisplaySkip

\newcommand\colorsquare[2][black]{\textcolor{#1}{\rule{#2}{#2}}}

\makeatother

\begin{document}

\title{Integrating Local Material Recognition\\
with Large-Scale Perceptual Attribute Discovery}

\author{Gabriel Schwartz\qquad{}\qquad{}Ko Nishino\\
Department of Computer Science, Drexel University\\
\texttt{\{gbs25,kon\}@drexel.edu}}
\maketitle

\begin{abstract}
Material attributes have been shown to provide a discriminative intermediate representation for recognizing materials, especially for the challenging task of recognition from local material appearance (i.e., regardless of object and scene context). In the past, however, material attributes have been recognized separately preceding category recognition. In contrast, neuroscience studies on material perception and computer vision research on object and place recognition have shown that attributes are produced as a by-product during the category recognition process. Does the same hold true for material attribute and category recognition? In this paper, we introduce a novel material category recognition network architecture to show that perceptual attributes can, in fact, be automatically discovered inside a local material recognition framework. The novel material-attribute-category convolutional neural network (MAC-CNN) produces perceptual material attributes from the intermediate pooling layers of an end-to-end trained category recognition network using an auxiliary loss function that encodes human material perception. To train this model, we introduce a novel large-scale database of local material appearance organized under a canonical material category taxonomy and careful image patch extraction that avoids unwanted object and scene context. We show that the discovered attributes correspond well with semantically-meaningful visual material traits via Boolean algebra, and enable recognition of previously unseen material categories given only a few examples. These results have strong implications in how perceptually meaningful attributes can be learned in other recognition tasks.
\end{abstract}

\section{\label{sec:Introduction}Introduction}

\begin{figure}[tb]
\begin{centering}
\includegraphics[width=0.98\columnwidth]{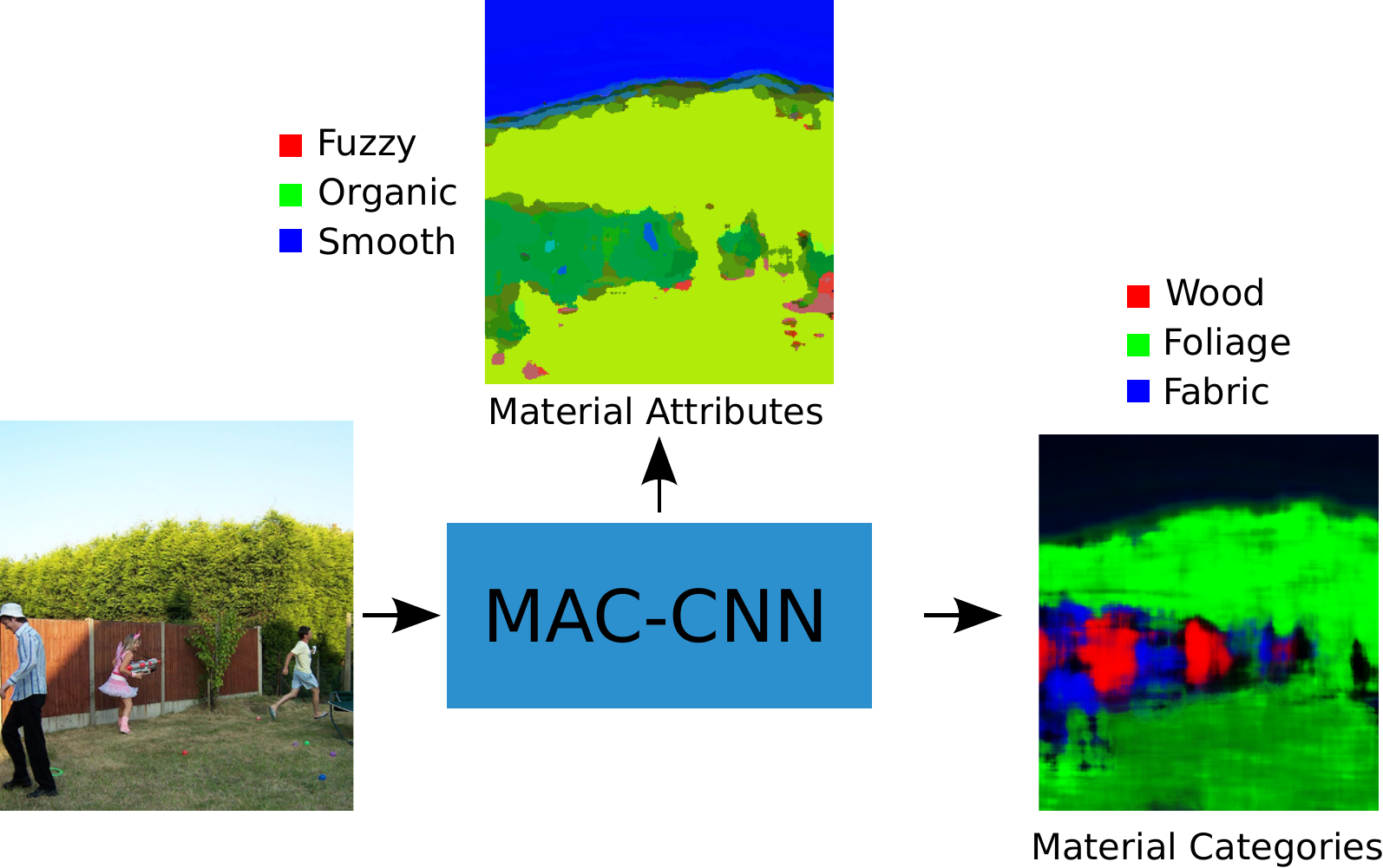}
\par\end{centering}
\caption{\label{fig:attrs_from_matrec}Existing attempts to leverage attributes
    for material recognition have recognized them separately from material categories. These approaches are inherently incompatible with
    large-scale recognition, from which attributes may be extracted, as semantic annotation of material attributes is challenging. In this paper, we show that we can
    automatically discover discriminative and semantically meaningful, perceptually-motivated, material
    attributes inside a local material recognition network trained end-to-end
    for category recognition.}
\end{figure}

Attributes have proven to be a valuable intermediate representation for
higher-level image understanding tasks. Material attributes, attributes that encode unique visual and non-visual material properties, are particularly useful as they provide a
discriminative representation for materials whose appearance otherwise exhibits
large intra-class variation~\cite{Schwartz2013}. Beyond just suggesting the
presence of various materials, material attributes can inform us as to the
potential physical properties, such as ``rough'' or ``soft'', a material might
exhibit. These cues can, for instance, guide autonomous interaction with
real-world surfaces made of various materials. Attributes, in general, also have the
desirable property that they can form a compact representation for unseen
categories from few examples (N-shot learning).

Existing material category and attribute recognition methods consider
attributes separately from category recognition. Attributes are used either
solely as an intermediate representation~\cite{Schwartz2013}, or as an
automatically discovered perceptual representation for the same
purpose~\cite{Schwartz2015,Vittayakorn2016}. In other words, material categories
are  defined on top of separately recognized material attributes.  As a result,
if both attributes and material recognition are required, images must pass
through two separate processes.

We would like to take advantage of the benefits of end-to-end learning to
incorporate automatically-discovered attributes with material recognition in
one seamless process. Material attribute recognition, however, is not easily
scalable. Past approaches rely on semantic attributes, such as ``shiny'' or
``fuzzy'', that need careful annotation by a consistent annotator as their
appearance may not be readily agreed upon. This precludes the use of large-scale crowdsourcing. We are also specifically interested
in local material recognition: recognizing materials using only information
from small patches inside object boundaries so as to separate materials from
the surrounding objects. This enables the recognition of material regardless of its situational context (e.g., what it makes up as an object or the place in which it is found), which is essential for realizing recognition of materials in general context (e.g., recognizing ceramic without knowing that the object is a cup).
Material recognition methods that rely on context like object shape fundamentally confuse objects
and materials: when that context is not available, their accuracy suffers~\cite{Schwartz2013,Schwartz2015}.

In this paper, we realize large-scale end-to-end learning for local material
recognition and show that perceptual material attributes (e.g., ``smooth'' and ``shiny'') can be extracted from the same framework. As depicted in Fig.~\ref{fig:attrs_from_matrec}, We introduce a
novel material attribute-category CNN architecture (MAC-CNN) to show that perceptual material
attributes, recognizable at the local level, can be discovered during material
recognition. By introducing additional auxiliary attribute layers (layers
connected to the network but not participating in the classification loss) and
constraints derived from human material perception, we find that we may
discover perceptual material attributes inside a material recognition
framework. Unlike methods that rely on images and text (along with material
annotations), we require weak supervision consisting only of a perceptual distance matrix of material categories to discover the attributes.

As part of our work, we also introduce a novel local material image database.
Despite the importance of local material recognition as demonstrated in~\cite{Schwartz2013,Schwartz2015}, existing material databases have been tailored for global material recognition based on large image patches or whole images that inevitably mix object appearance with material appearance.
Local image patches can be extracted from the Flickr Materials Database~\cite{Sharan2009}, but the use of only Flickr images biases the dataset towards more artistic or professional images. Recent datasets, such as the Materials in Context (MINC) dataset~\cite{Bell2015}, take steps to address this, but have inconsistencies in the definition of what makes a material category (e.g., ``mirror'' and ``carpet'' which are obviously objects are used as materials). The patches they extract are also large enough to include entire objects, further confusing the recognition of objects and materials. Fig. 10 of \cite{Bell2015} clearly shows that objects are recognized to identify materials (e.g., ``mirror'' as a material is recognized by finding actual mirrors and ``fabric'' is recognized by finding pillows). In contrast, we introduce the first comprehensive large-scale database explicitly targeted at local material recognition. We derive a systematically organized hierarchy for material categories, and we collect annotations for images from a wide variety of sources while carefully ensuring that object information, such as shape, is not present.

Interesting parallels can be found in recent neuroscience studies that reveal that human material perception
produces an internal representation corresponding to semantic material
attributes. Hiramatsu~\etal\cite{Hiramatsu2011} and Goda~\etal\cite{Goda2014} have
investigated how visual information is transformed in the brain during the
human and primate recognition of materials. They find that the material
representation in our visual system shifts from raw image features at lower
levels (V1/V2) to perceptual properties (such as matte, colorful, fuzzy, shiny,
etc.) in higher-level brain regions dedicated to recognition (FG/CoS). On the other hand, in computer vision, specifically in the
separate domain of conventional object recognition, Zhou~\etal\cite{Zhou2015_2}
find that object detectors appear in scene recognition CNNs. Our work serves as
further support for the idea of semantically meaningful attributes arising inside category recognition process, by showing that inherent material attributes can be made explicit inside a material recognition process.

Our results show that MAC-CNN produces a generalizable internal material
representation.
We show that the attributes we extract exhibit the same
properties, such as spatial consistency, as existing automatically-discovered
perceptual material attributes. By visualizing the arrangement of material
categories in the space of attribute probabilities, we show that attributes
separate materials into distinct clusters. We perform true local material
recognition, predicting categories for single small image patches with no
aggregation, a significantly more challenging task than previous approaches.
While previous work suggests that perceptual attributes are correlated with
manually-identified semantic material traits, such as ``fuzzy'' or ``smooth'',
we are the first to conclusively demonstrate this by recognizing them solely
from our extracted attributes with logic regression. Finally, we demonstrate that the extracted material
attributes add significant information to recognize previously unseen material
categories from a small number of training examples (i.e., N-shot learning with
material attributes). These results show that our method successfully extracts
effective and semantically meaningful internal representations of complex
material appearance from a local material recognition network. These results also suggest a general approach for extracting semantically meaningful, perceptually-motivated attributes in general recognition processes, such as object and place recognition.

\section{\label{sec:Related-Work}Related Work}

In this paper, we investigate convolutional neural networks (CNNs) as the
framework within which we should find perceptual material attributes. In the past, for recognition tasks other than materials such as object and places, research on localizing attributes inside CNNs for category recognition has been explored. 
Specifically for object attributes and categories, Shankar~\etal\cite{Shankar2015} recently proposed a modified training
procedure called ``deep carving'' which
provides the CNN with attribute pseudo-label targets, updated periodically
during training. This causes the resulting network to be better-suited for
object attribute prediction. Escorcia~\etal\cite{Escorcia2015} showed that known
semantic object attributes can be extracted from a CNN. Similar to our work,
they showed that object attributes depend on features in all layers of the CNN.
ConceptLearner, proposed by Zhou~\etal\cite{Zhou2015} uses weak supervision, in
the form of images with associated text content, to discover semantic
place attributes that can be interpreted as object descriptions. These attributes correspond to terms within the text that appear in
the images. All of these frameworks predict a single set of object or place attributes for an
entire image, as opposed to the per-pixel material attributes discussed in our work.
Furthermore, our extracted  attributes do not require semantic information
(which may be challenging to collect in a consistent manner), and are defined
based on human perceptual information.

At the intersection of neuroscience and computer vision,
Yamins~\etal\cite{Yamins2014} find that feature responses from high-performing
CNNs can accurately model the neural response of the human visual system in the
inferior temporal (IT) cortex. They perform a linear regression from CNN feature
outputs to IT neural response measurements and find that the CNN features are
good predictors of neural responses. Their work focuses on object
recognition CNNs, not materials. Hiramatsu~\etal\cite{Hiramatsu2011} take
functional magnetic resonance imaging (fMRI) measurements and investigate their
correlation with both direct visual information and perceptual material
properties (similar to the material traits of~\cite{Schwartz2013}) at various
areas of the human visual system. They find that pairwise material
dissimilarities derived from fMRI data correlate best with direct visual
information (analogous to pixels) at the lower-order areas and with perceptual
attributes at higher-order areas. Goda~\etal\cite{Goda2014} obtain similar
findings in non-human primates. Of particular importance is the fact that their
work inherently considers materials independently from objects. Material
samples are shown with the same cylinder shapes, thus avoiding any distracting object cues. These studies suggest the existence of
perceptual material attributes in human local material recognition.

Our work is closely related to the non-semantic perceptual material attributes
discovered by Schwartz and Nishino~\cite{Schwartz2015}. In their work, they
collect measurements of human perceptual distances between material categories
and use those distances to discover perceptual material attributes that
reproduce these distances. These attributes are subsequently used to recognize
material categories. We use the constraints derived in their work as a basis
for our auxiliary attribute layers. This approach can be considered similar to
the work of Lee~\etal\cite{Lee2015}, which introduced ``deep supervision'' via
auxiliary loss functions to better-propagate gradient information during CNN
training for object recognition. They do so by adding additional SVM-like loss functions that
encourage classification at lower levels of the network. Rather than
simply replicating the final classification loss, we impose new constraints to
explicitly output additional information about the input, in our case the
perceptual material attributes.

\section{\label{sec:Discovering-Perceptual-Attribute}Perceptual Material
Attributes from Local Material Recognition}

In this paper, we show that perceptual material attributes can be integrated
with a local material recognition framework and output as a side-product. We
find the human-perception-based attributes of Schwartz and
Nishino~\cite{Schwartz2015} to be particularly relevant, as they automatically
discover material attributes from weak supervision. Their attributes are,
however, recognized separately from materials in a slow process that scales
poorly with more training data. In this section we derive a novel framework to
discover perceptual attributes similar to those in~\cite{Schwartz2015}, inside
a CNN framework, while simultaneously learning to classify materials.

A straightforward approach to integrating material attributes and category
recognition would be to add an attribute prediction layer at the top of a
material recognition CNN, immediately before the final material category
probability softmax layer. As an initial investigation, we implemented this
approach with the goal of predicting the perceptual attributes derived
from~\cite{Schwartz2015}. We, however, found that constraining the network in such a fashion results in either poorly-recognized
attributes or categories.

These results suggest that materials are not defined simply by their
attributes. This agrees with the findings of
Hiramatsu~\etal\cite{Hiramatsu2011}, where they note that the human neural
representation of material categories transitions from visual (raw image
features) to perceptual (visual properties like ``shiny'') in an hierarchical
fashion. This also suggests that material attributes require information from multiple levels of the material recognition network.

\subsection{\label{subsec:Combined-Attribute/Material-CNN}Material
Attribute-Category CNN}

\begin{figure*}[tb]
\begin{centering}
\includegraphics[width=1.6\columnwidth]{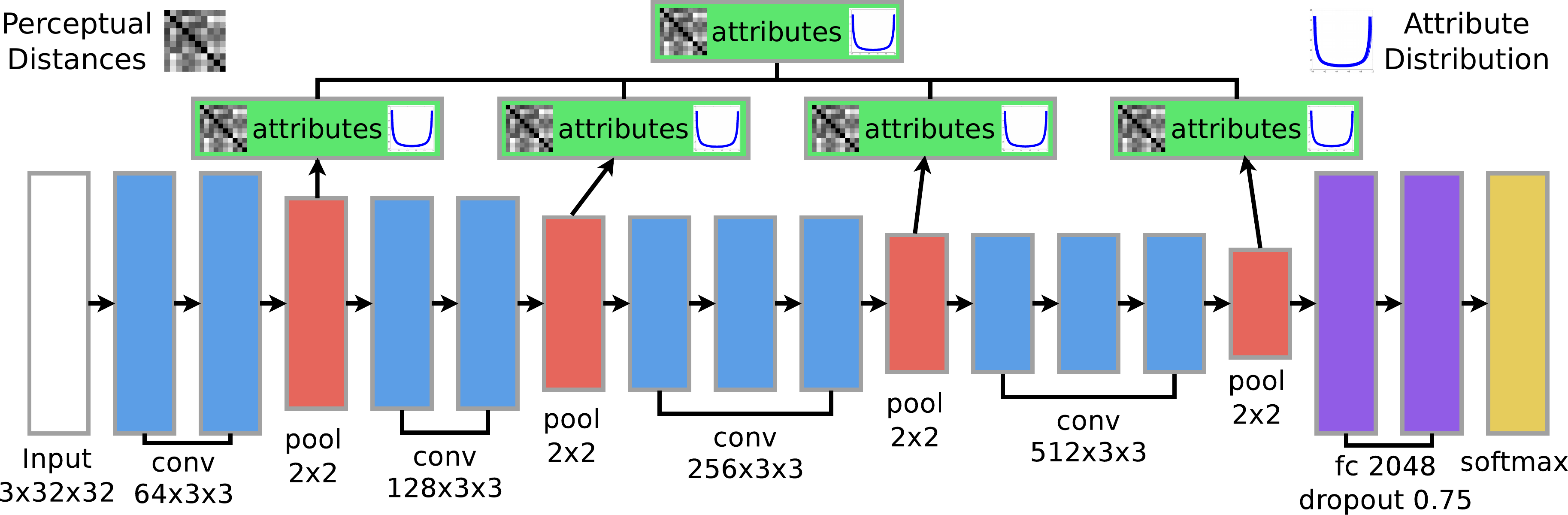}
\par\end{centering}
\caption{\label{fig:CNN-Architecture}Material Attribute-Category CNN (MAC-CNN)
Architecture: We introduce auxiliary fully-connected attribute layers
to each spatial pooling layer and combine the per-layer predictions
into a final attribute output via an additional set of weights. The
loss functions attached to the attribute layers encourage the extraction
of attributes that match the human material representation encoded
in perceptual distances. The first set of attribute layers acts as
a set of weak learners to extract attributes wherever they are present.
The final layer combines them to form a single prediction.}
\end{figure*}

We need a means of extracting attribute information at multiple levels of the
network. Simply combining all feature maps from all network layers and using
them to predict attributes would be computationally impractical.  Rather than
directly using all features at once, we augment an initial CNN designed for
material classification with a set of auxiliary fully-connected layers attached
to the spatial pooling layers. This allows the attribute layers to use
information from multiple levels of the network without needing direct access
to every feature map. We treat the additional layers as a set of weak learners,
each auxiliary layer discovering the attributes available at the corresponding
level of the network.  This concept is similar to deep supervision by
Lee~\etal\cite{Lee2015}.  Their goal, however, is to inject the category
recognition loss function into intermediate layers for better end recognition
(in their case, object recognition) by simply propagating the same
classification targets (via SVM-like loss functions) to the lower layers. Our
goal is to discover and extract perceptual material attributes through this
internal supervision using loss functions different from that for material
category recognition. 

For the auxiliary layer loss functions, we extend the perceptual attribute loss
functions of~\cite{Schwartz2015} and apply them to the outputs of each
auxiliary fully-connected layer. Schwartz and Nishino's proposed method begins
with a set of pairwise perceptual distances between material categories
measured via human yes/no binary similarity annotations on material image
patches. From these distances, they learn a mapping matrix $\mathbf{A}$ between
categories and unknown, non-semantic attributes. The mapping preserves the
pairwise human perceptual distances while causing the resulting attributes to
exhibit the behaviors, such as spatial consistency, of semantic attributes.  We
derive our attribute layer loss functions from these learning constraints.

Specifically, assuming the output of a given pooling layer $i$ in
the network for image $j$ is $\mathbf{h}_{ij}$, and given categories
$C,\,\left|C\right|=K$ and a set of sample points $P\in\left(0,1\right)$
for density estimation, we add these auxiliary loss functions:
\begin{equation}
u_{i}=\frac{1}{K}\sum_{k\in C}\left\Vert \mathbf{a}_{k}-\frac{1}{N_{k}}\sum_{j|c_{j}=k}f\left(\mathbf{W}_{i}^{\mathrm{T}}\mathbf{h}_{ij}+\mathbf{b}_{i}\right)\right\Vert _{1}\label{eq:unary}
\end{equation}
\begin{equation}
d_{i}=\sum_{p\in P}\beta\left(p;a,b\right)\ln\left(\frac{\beta\left(p;a,b\right)}{q\left(p;f\left(\mathbf{W}_{i}^{\mathrm{T}}\mathbf{h}_{ij}+\mathbf{b}_{i}\right)\right)}\right)\text{,}\label{eq:distr}
\end{equation}
where $f\left(x\right)=\min\left(\max\left(x,0\right),1\right)$ clamps
the outputs within $\left(0,1\right)$ to conform to attribute probabilities,
and weights $\mathbf{W}_{i},\mathbf{b}_{i}$ represent the auxiliary
fully-connected layers we add to the network. $\mathbf{a}_{k}$ represents
a row in the category-attribute mapping matrix we derived from our
data by collecting the yes/no similarity answers used in~\cite{Schwartz2015}
for patches in our database (see Sec.~\ref{sec:Constructing-a-Local-Material-Database}).
Equation~\ref{eq:unary} causes the attribute layer to discover attributes
which match the perceptual distances measured from human annotations.
As certain attributes are expected to appear at different levels of
the network, some layers will be unable to extract them. This implies
that their error should be sparse, either predicting an attribute
well or not at all. For this reason we use an L1 error norm. Equation~\ref{eq:distr},
applied only to the final attribute layer, encourages the distribution
of the attributes to match those of known semantic material traits.
It takes the form of a KL-divergence between a Beta distribution (empirically
observed by~\cite{Schwartz2015} to match the distribution of semantic
attribute probabilities), and a Kernel Density Estimate $q\left(\cdot\right)$
of the extracted attribute probability sampled at points $p\in P$.

The reference network we build on is based on the high-performing VGG-16
network of Simonyan and Zisserman~\cite{Simonyan2015}. We use their trained
convolutional weights as initialization where applicable, and add new
fully-connected layers for material classification.
Fig.~\ref{fig:CNN-Architecture} shows our architecture for material attribute
discovery and category recognition. We refer to this network as the Material
Attribute-Category CNN (MAC-CNN).

\section{\label{sec:Constructing-a-Local-Material-Database}Local Material
Database}

In order to train the category recognition portion of the MAC-CNN, we need a
proper local material recognition dataset. We find existing material databases
lacking in a few key areas necessary to properly perform local material
recognition. Previous material recognition datasets
\cite{Sharan2013,Bell2013,Bell2015} have relied on ad-hoc choices regarding the
selection and granularity of material categories (e.g., carpet and wall are
considered materials).  When patches are involved, as in~\cite{Bell2015}, the
patches can be as large as 24\% of the image size surrounding a single pixel
identified as corresponding to a material. These patches are large enough to
include entire objects. These issues make it difficult to separate challenges
inherent to material recognition from those related to general recognition
tasks and inevitably lead to material recognition based on object and scene
information, which would not be beneficial for scene understanding tasks (e.g., recognized material information will not help recognize objects and places as it already relies on the recognition of them).  We
also find that image diversity is still lacking in modern datasets: FMD~\cite{Sharan2013} is solely sourced from Flickr which is heavily biased towards professional photography and MINC~\cite{Bell2013} is predominantly sourced from professional real-estate photography. We introduce a new local material recognition dataset to support the
experiments in this paper.

\subsection{\label{subsec:Material-Hierarchy}Material Category Hierarchy}

\begin{figure}[tb]
\begin{centering}
\includegraphics[width=0.98\columnwidth]{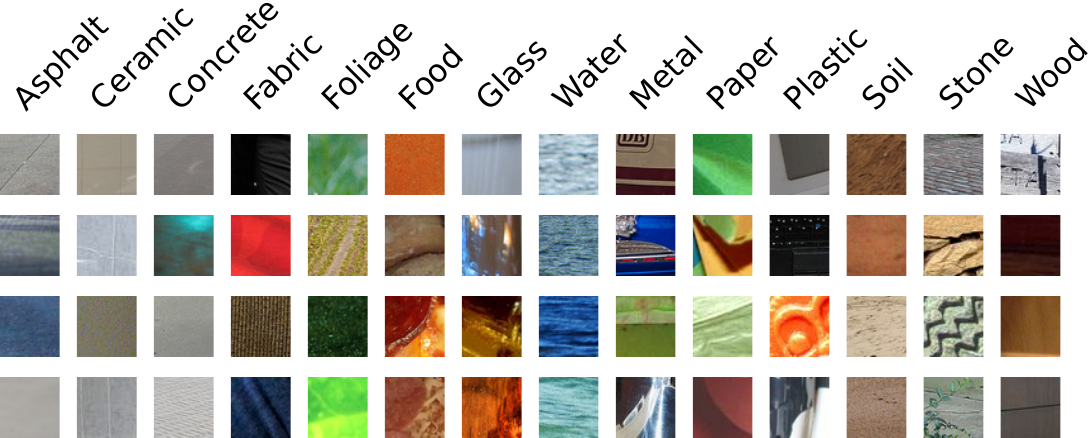}
\par\end{centering}
\caption{\label{fig:Local-Material-Patches}Local material patches extracted as
    the final step in our database creation process. These patches are used to
    compute human perceptual distances, and also form the training input for
    our combined material attribute-category CNN.}

\end{figure}
Material categories in existing datasets have been selected in a rather adhoc manner in the past.
Examples of this issue include the proposed material categories ``mirror''
(actually an object), and ``brick'' (an object or group of objects).  Existing
categories also confuse materials and their properties (e.g., surface finish),
for example, separating ``stone'' from ``polished stone''. To address the issue
of material category definition, we propose a more carefully-selected set of
material categories for local material recognition. We derive a taxonomy of
materials based on canonical categorization in materials science~\cite{matbase} and
create a hierarchy based on the generality of each material family. Please see 
our supplemental material for a complete diagram of the hierarchy including all
categories at all levels.

Our hierarchy consists of a set of three-level material trees. The highest
level corresponds to major structural differences between materials in the
category. Metals are conductive, polymers are composed of long chain molecules,
ceramics have a crystalline structure, and composites are fusions of materials
either bonded together or in a matrix. We define the mid-level (whcih can also be referred
to as entry-level~\cite{Ordonez2013}) categories as groups that separate
materials based primarily on their visual properties. Rubber and paper are
flexible, for example, but paper is generally matte and rubber exhibits little
color variation.  The lowest level, fine-grained categories, can often only be
distinguished via a combination of physical and visual properties. Silver and
steel, for example, may be challenging to distinguish based solely on visual
information.

Such a hierarchy is sufficient to cover most natural and manmade materials.  In
creating our hierarchy, however, we found that certain categories that are in
fact materials did not fit within the strict definitions described above. For
the sake of completeness, we make the conscious decision to add these mid-level
categories to our data collection process. These categories are: food, water,
and non-water liquids.  While food is both a material and an object, we rely on
our annotation process (Sec.~\ref{subsec:Crowdsourcing-Annotation}) to ensure
we obtain examples of the former and not the latter.

\subsection{\label{subsec:Crowdsourcing-Annotation}Data Collection and
Annotation}

\begin{figure}[tb]
\begin{centering}
\includegraphics[width=0.98\columnwidth]{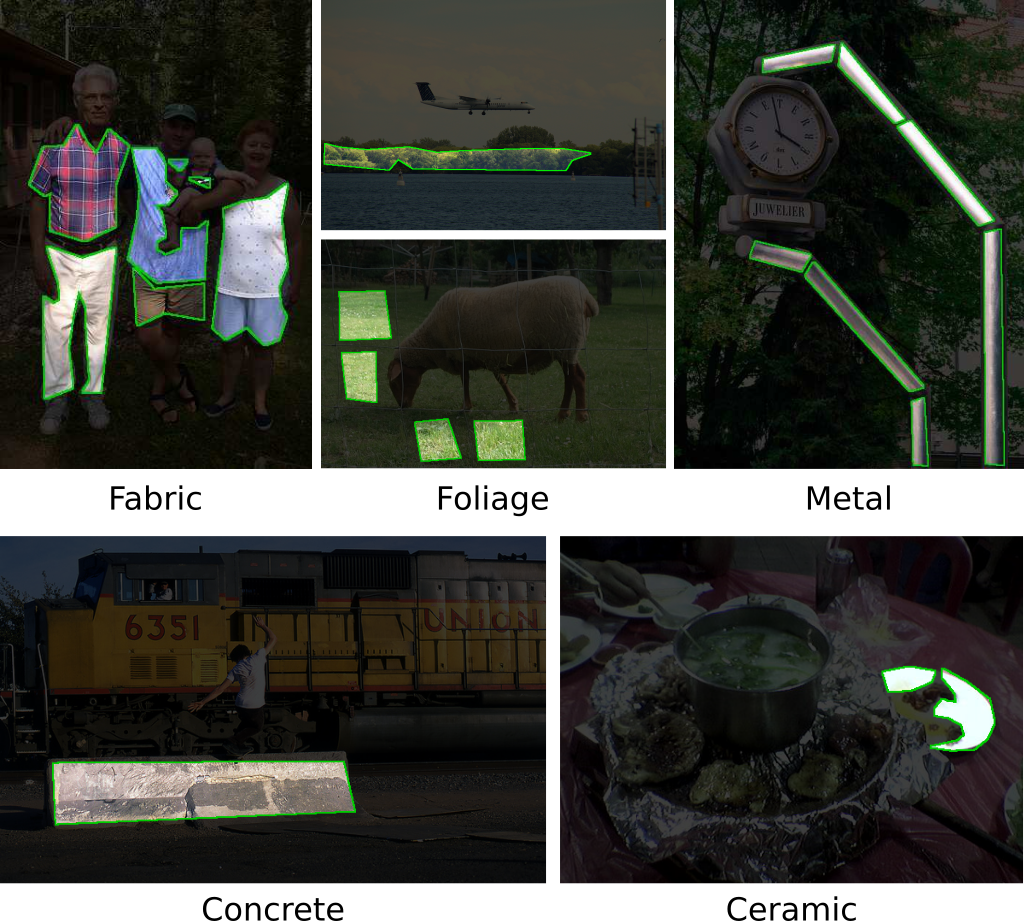}
\par\end{centering}
\caption{\label{fig:annotation-results}Annotators did not hesitate to take
    advantage of the ability to draw multiple regions, and most understood the
    guidelines concerning regions crossing object boundaries. As a result, we
    have a rich database of segmented local material regions.}
\end{figure}

The mid-level set of categories forms the basis for a crowdsourced annotation
pipeline to obtain material regions from which we may extract local material
patches (Fig.~\ref{fig:Local-Material-Patches}).  We employ a multi-stage
process to efficiently extract both material presence and segmentation
information for a set of images.
The first stage asks annotators to identify materials present in the image.
Given a set of images with materials identified in each image, the second stage
presents annotators with a user interface that allows them to draw multiple
regions in an image. Each annotator is given a single image-material pair and
asked to mark regions where that material is present. While not required, our
interface allows users to create and modify multiple disjoint regions in a
single image.  Images undergo a final validation step to ensure no poorly drawn
or incorrect regions are included.

Each image in the first stage is shown to multiple annotators and a consensus
is taken to filter out unclear or incorrect identifications.  While sentinels
and validation were not used to collect segmentations in other datasets, ours
is intended for local material recognition.  This implies that identified
regions should contain only the material of interest. During collection,
annotators are given instructions to keep regions within object boundaries, and
we validate the final image regions to insure this.

Image diversity is an issue present to varying degrees in current material
image datasets. The Flickr Materials Database (FMD)~\cite{Sharan2009} contains
images from Flickr which, due to the nature of the website, are generally more
artistic in nature. The OpenSurfaces and Materials in Context
datasets~\cite{Bell2013,Bell2015} attempt to address this, but still draw from
a limited variety of sources (e.g., real estate photographs). We source our
images from multiple existing image datasets spanning the space of indoor,
outdoor, professional, and amateur photographs. We use images from the PASCAL
VOC database~\cite{Everingham2010}, the Microsoft COCO
database~\cite{Lin2014}, the FMD~\cite{Sharan2009}, and the ImageNet
database~\cite{Russakovski2015}.
\begin{figure}[tb]
\begin{centering}
\includegraphics[width=0.98\columnwidth]{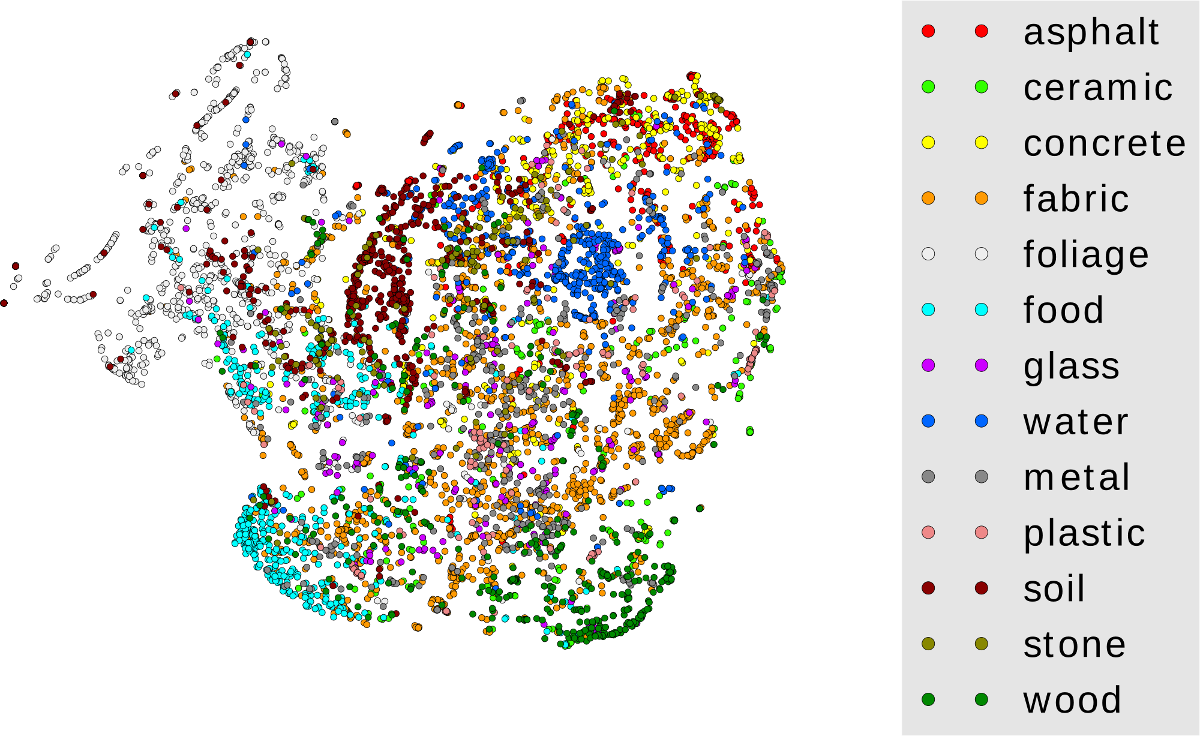}
\par\end{centering}
\caption{\label{fig:Attribute-Analysis}Attribute Space Embedding via
    t-SNE~\cite{Hinton2008}: Many categories, such as water, food, foliage,
    soil, and wood, are very well-separated in the attribute space. We find
    that this separation corresponds roughly with per-category accuracy.}
\end{figure}

Examples in Fig.~\ref{fig:annotation-results} show that our annotation
pipeline successfully provides properly-segmented material regions within many
images. Many images also contain multiple regions. While the level of detail
for provided regions varies from simple polygons to detailed material
boundaries, the regions all contain single materials. The final database contains 2669 images with associated material segmentations. We may extract at least 200,000 image patches of decent size (e.g., $48\times 48$) from inside the segmented regions without crossing object boundaries from this database. The database and the code for MAC-CNN will be made publicly accessible after publication.

\section{\label{sec:Analysis-of-Perceptual}Perceptual Attributes in the MAC-CNN}
\begin{figure}[tb]
\begin{centering}
\includegraphics[width=0.98\columnwidth]{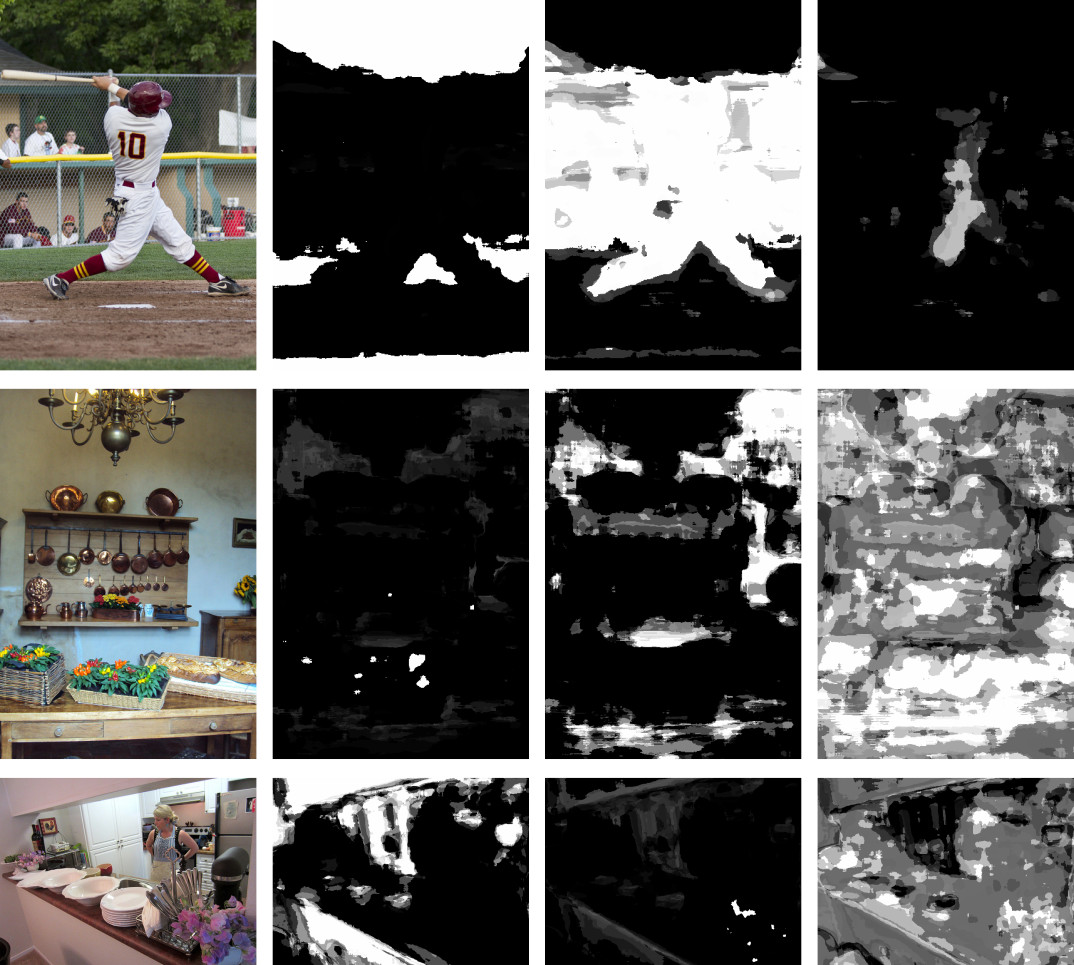}
\par\end{centering}
\caption{\label{fig:Per-Pixel-Attribute-Probabilitie}Each column after the
    first (the input image) shows per-pixel probabilities for an extracted
    perceptual attribute. The attributes form clearly delineated regions,
    similar to semantic attributes, and their distributions match as well.}
\end{figure}

To verify that the perceptual attributes we seek can in fact be extracted with
our MAC-CNN, we augment our dataset with annotations to compute the necessary
perceptual distances described in~\cite{Schwartz2015}.  Using our dataset and
these distances, we derive a category-attribute matrix $\mathbf{A}$ and train
an implementation of the MAC-CNN described in
Sec.~\ref{subsec:Combined-Attribute/Material-CNN}.

We train the network on \textasciitilde{}200,000 $48\times48$ image patches
extracted from segmented material regions. Optimization is performed using
mini-batch stochastic gradient descent with momentum.  The learning rate is
decreased by a factor of 10 whenever the validation error increases, until the
learning rate falls below $1\times10^{-8}$.

\subsection{Perceptual Material Attribute Properties}

\begin{figure}[tb]
\begin{centering}
\begin{minipage}[b][1\totalheight][t]{0.45\columnwidth}%
\begin{center}
\includegraphics[width=0.48\columnwidth]{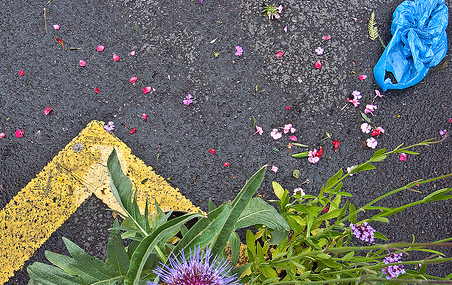}
\includegraphics[width=0.48\columnwidth]{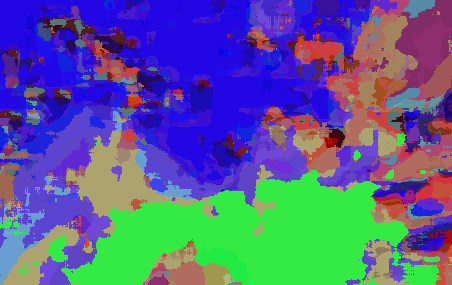}\vspace{-27pt}
\par\end{center}
\begin{center}
\begin{tabular}{>{\centering}p{0.36\columnwidth}>{\centering}p{0.3\columnwidth}>{\centering}p{0.3\columnwidth}}
\colorsquare[red]{4pt}\textsf{\scriptsize{} Manmade} & \colorsquare[green]{4pt}\textsf{\scriptsize{} Organic} & \colorsquare[blue]{4pt}\textsf{\scriptsize{} Rough}\tabularnewline
\end{tabular}\vspace{-5pt}
\par\end{center}
\begin{center}
\includegraphics[width=0.48\columnwidth]{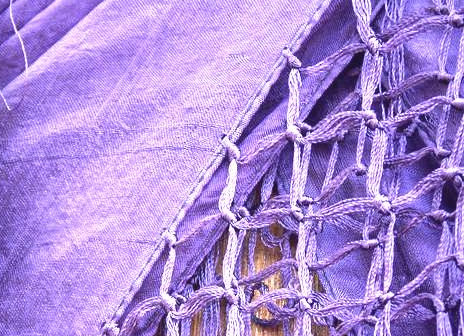}
\includegraphics[width=0.48\columnwidth]{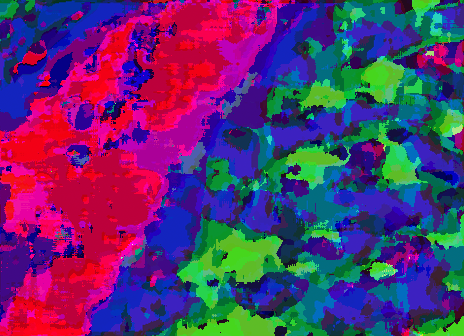}\vspace{-15pt}
\par\end{center}
\begin{center}
\begin{tabular}{ccc}
\colorsquare[red]{4pt}\textsf{\scriptsize{} Smooth} & \colorsquare[green]{4pt}\textsf{\scriptsize{} Striped} & \colorsquare[blue]{4pt}\textsf{\scriptsize{} Soft}\tabularnewline
\end{tabular}\vspace{-5pt}
\par\end{center}
\begin{center}
\includegraphics[width=0.48\columnwidth]{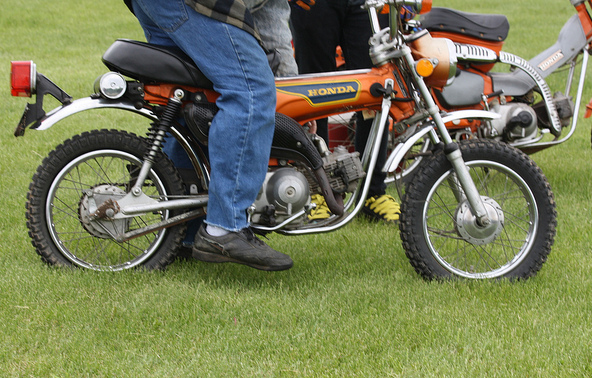}
\vspace{-15pt}
\includegraphics[width=0.48\columnwidth]{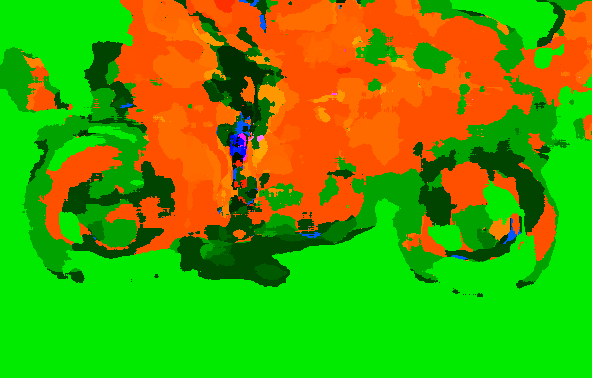}\vspace{-10pt}
\par\end{center}
\begin{center}
\begin{tabular}{>{\centering}p{0.33\columnwidth}>{\centering}p{0.33\columnwidth}>{\centering}p{0.33\columnwidth}}
\colorsquare[red]{4pt}\textsf{\scriptsize{} Metallic} & \colorsquare[green]{4pt}\textsf{\scriptsize{} Organic} & \colorsquare[blue]{4pt}\textsf{\scriptsize{} Smooth}\tabularnewline
\end{tabular}
\par\end{center}%
\end{minipage} \hspace{5pt}%
\begin{minipage}[b][1\totalheight][t]{0.49\columnwidth}%
\begin{center}
\includegraphics[width=0.48\columnwidth]{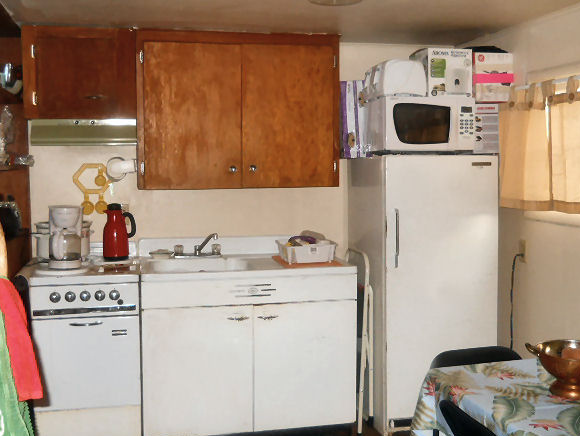}
\includegraphics[width=0.48\columnwidth]{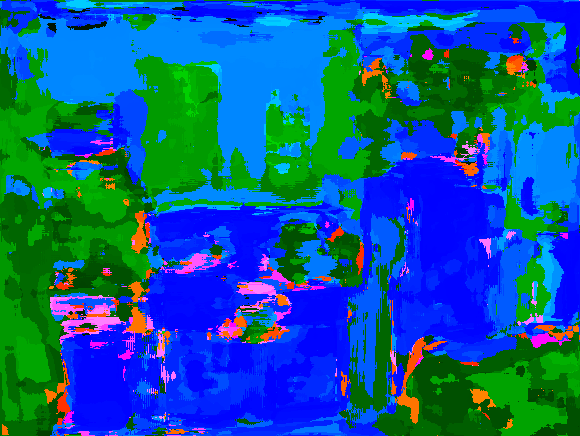}\vspace{-15pt}
\par\end{center}
\begin{center}
\begin{tabular}{ccc}
\colorsquare[red]{4pt}\hspace{3pt}\textsf{\scriptsize{}Shiny\hspace{5pt}} & \colorsquare[green]{4pt}\hspace{3pt}\textsf{\scriptsize{}Organic\hspace{5pt}} & \colorsquare[blue]{4pt}\hspace{3pt}\textsf{\scriptsize{}Smooth}\tabularnewline
\end{tabular}
\par\end{center}
\vspace{0pt}

\begin{center}
\includegraphics[width=0.49\columnwidth]{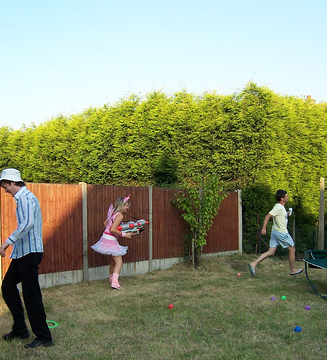}
\includegraphics[width=0.49\columnwidth]{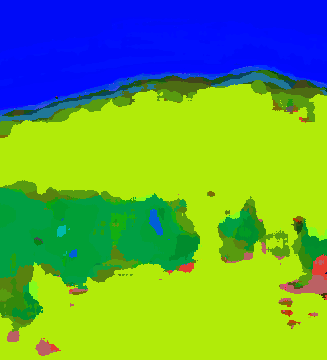}\vspace{-15pt}
\par\end{center}
\begin{center}
\begin{tabular}{ccc}
\colorsquare[red]{4pt}\hspace{3pt}\textsf{\scriptsize{}Fuzzy} & \colorsquare[green]{4pt}\hspace{3pt}\textsf{\scriptsize{}Organic} & \colorsquare[blue]{4pt}\textsf{\scriptsize{}Smooth}\tabularnewline
\end{tabular}
\par\end{center}%
\end{minipage}
\par\end{centering}
\caption{\label{fig:traits-from-attrs}By performing logic regression from our
    MAC-CNN extracted attributes to material traits, we are able to extract
    semantic information from our non-semantic attributes.  Doing so in a
    sliding window gives per-pixel semantic material trait information. The
    predictions show crisp regions that correspond well with their associated
    semantic traits. Traits are independent, and thus the maps contain mixed
    colors. Fuzzy and organic in the lower right image, for example, creates a
    yellow tint. These semantic material traits computed from discovered
    material attributes provide rich information about the underlying surface
    properties that can be leveraged to determine how to interact with them.}
    \end{figure}

We examine the properties of our perceptual material attributes by visualizing
how they separate materials, computing per-pixel attribute maps to verify that
the attributes are being recognized consistently, and linking the non-semantic
attributes with known semantic material traits (``fuzzy'', ``smooth'', etc...)
to visualize semantic content.  Figs.~\ref{fig:Attribute-Analysis},
\ref{fig:Per-Pixel-Attribute-Probabilitie}, and~\ref{fig:traits-from-attrs} are
generated using a test set of held-out images.

A 2D embedding of material image patches shows that the perceptual attributes
(Fig.~\ref{fig:Attribute-Analysis}) separate material categories. A number of
materials are almost completely distinct in the attribute space, while a few
form overlapping but still distinguishable regions. Foliage, food, and water
form particularly clear clusters.  The quality of the clusters matches the
per-category recognition rates, with accurately-recognized categories forming
more separate clusters.

Visualizations of per-pixel attribute probabilities in
Fig.~\ref{fig:Per-Pixel-Attribute-Probabilitie} show that the attributes are
spatially consistent. While overfitting is difficult to measure for
weakly-supervised attributes, we use spatial consistency as a proxy. Spatial
consistency is an indicator that the attributes are not overly-sensitive to
minute changes in local appearance, something that would appear if overfitting
were present. The attributes exhibit correlation with the materials that
induced them: attributes with a strong presence in a material region in one
image often appear similarly in others. The visualizations also clearly show
that the attributes are representing more than trivial properties such as
``flat color'' or ``textured''.

Logic regression~\cite{Ruczinski2003} is a method for building trees that
convert a set of boolean variables into a probability value via logical
operations (AND, OR, NOT). It is well-suited for collections of binary
attributes such as ours. Results of performing logic regression
(Fig.~\ref{fig:traits-from-attrs}) from extracted attribute predictions to
known semantic material traits (such as fuzzy, shiny, smooth etc...) show that
our MAC-CNN attributes encode material traits with the same average accuracy
(75\%) as the attributes of~\cite{Schwartz2015}.  For per-trait accuracy
comparisons, please see our supplemental materials.  We may also predict
per-pixel trait probabilities in a sliding window fashion, showing that the
attributes are encoding both perceptual and semantic material properties. The
material attributes provide rich information regarding the surface properties
that may benefit, for instance, action planning for autonomous agents.

\subsection{\label{subsec:Local-Material-Recognition}Local Material Recognition}

\setlength\tabcolsep{1.25pt} \def\arraystretch{0.5}

\begin{figure}[tb]
\begin{centering}
\includegraphics[width=0.98\columnwidth]{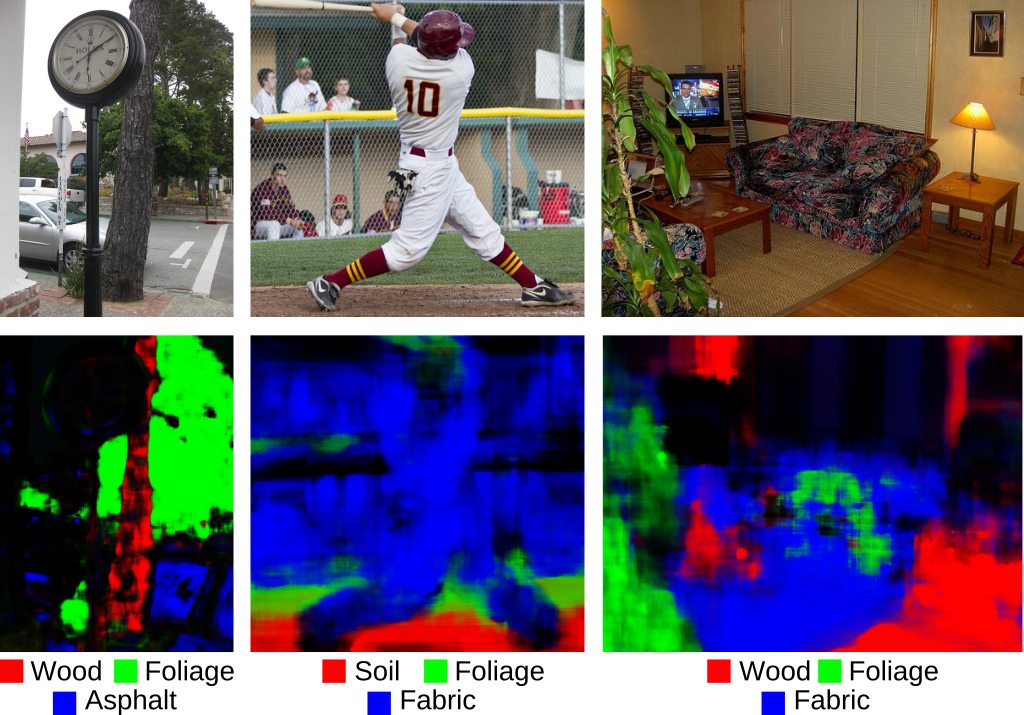}
\par\end{centering}
\caption{\label{fig:Material-Probability-Maps}These material maps, obtained by
    applying the MAC-CNN in a sliding window, show that we may obtain coherent
    regions using only small local patches as input. The foliage predictions on
    the couch are reasonable, as the local appearance pattern is indeed a
    flower. In the baseball image, the local appearance of the fence resembles
    lace (a fabric).}
\end{figure}

To evaluate the material recognition portion of the MAC-CNN output, we compute
local material recognition accuracy using the MAC-CNN trained on our database.
Accuracy is measured as the average number of correct patch category
predictions.  Average local material accuracy is 60.2\% across all categories.
Foliage is the most accurately recognized, consistent with past material
recognition results in which foliage is the most visually-distinct category.
Paper is the least well-recognized category. Unlike the artistic closeup 
images of the FMD, many of the images in our database come from ordinary 
images of scenes. Paper, in these situations, shares its appearance with a 
number of other materials such as fabric. These results can be viewed as a 
baseline accuracy for this dataset using a VGG architecture trained with small  
patches. It is important to note that we are  recognizing materials directly  
from single small image patches, with none of 
the region-based aggregation or large patches used 
in~\cite{Schwartz2013,Schwartz2015,Bell2015}. This is a much more challenging 
task as the available information is restricted. For a breakdown of 
per-category accuracy, please see our supplemental material.

\begin{figure*}[t]
\begin{centering}
\vspace{-18pt}
\includegraphics[width=0.66\columnwidth]{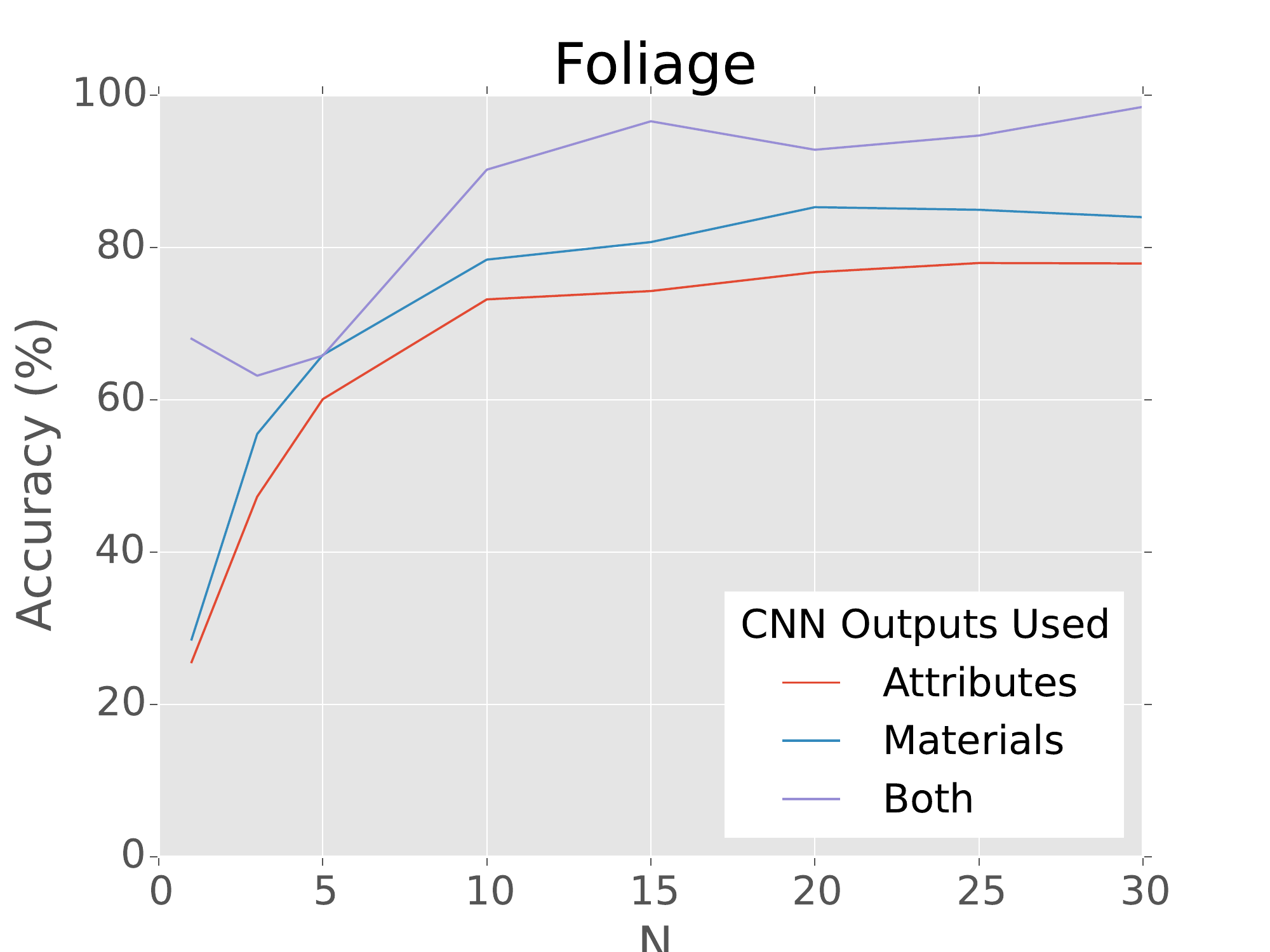}\includegraphics[width=0.66\columnwidth]{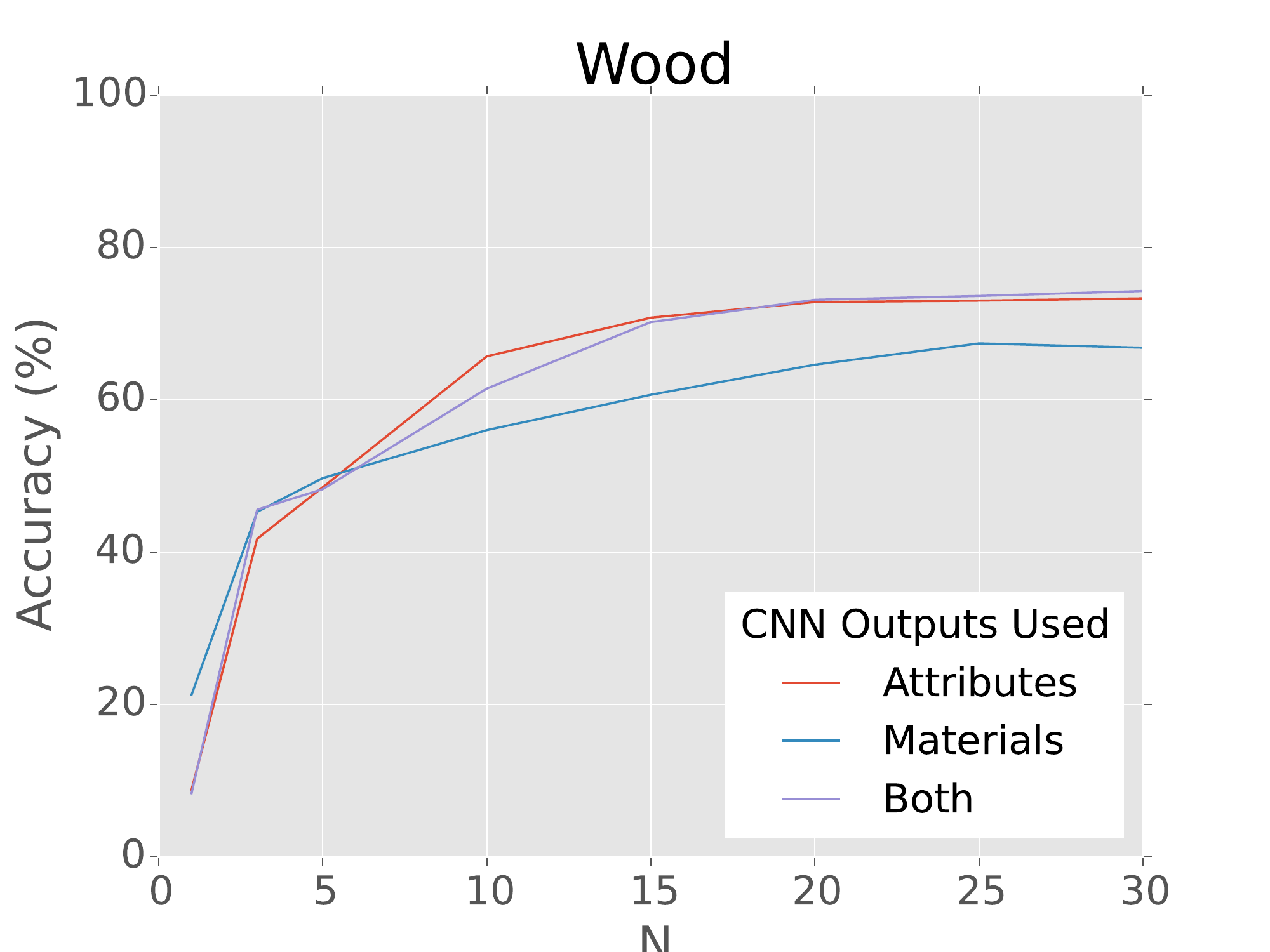}\includegraphics[width=0.66\columnwidth]{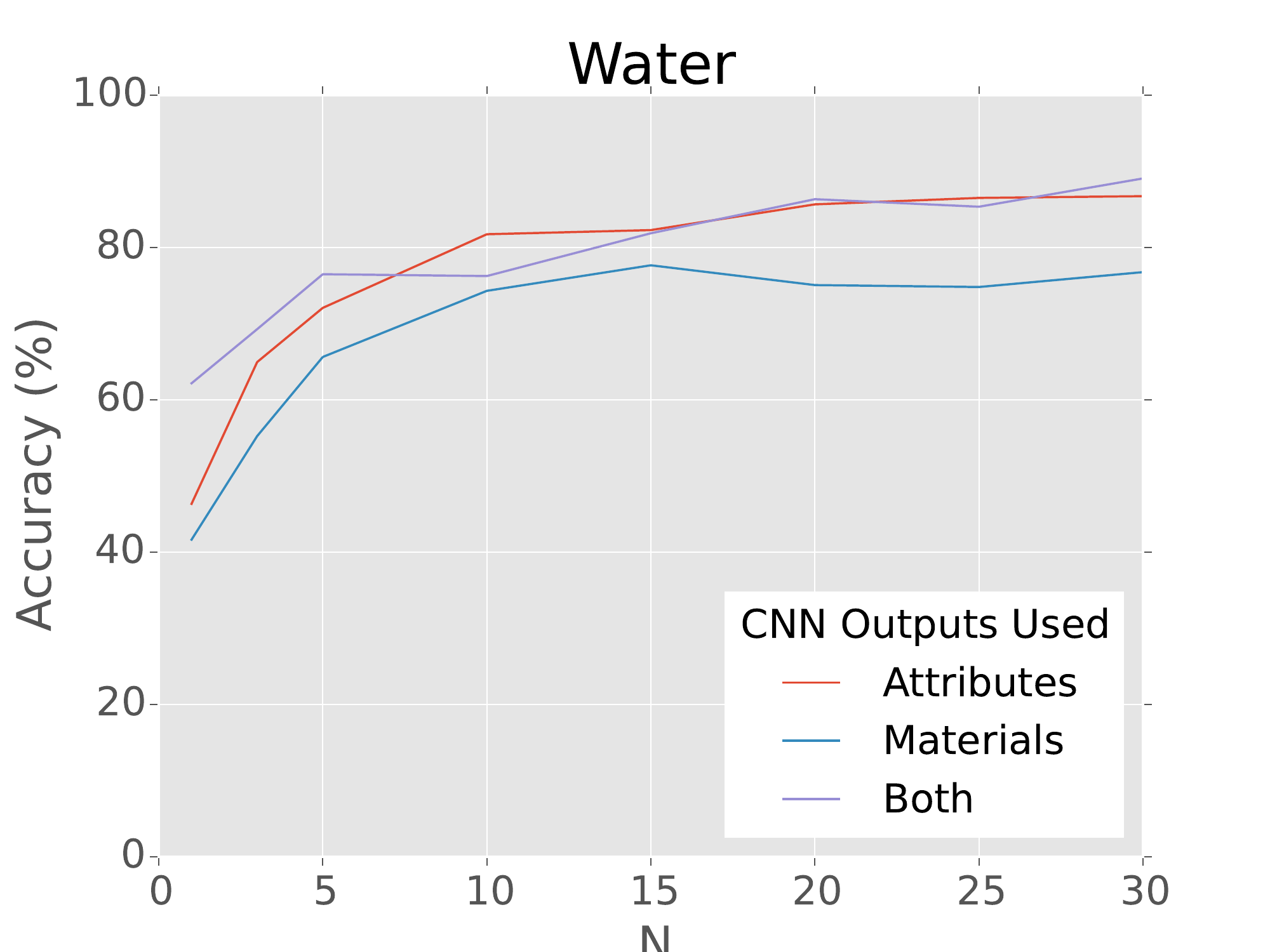}
\par\end{centering}
\caption{\label{fig:Transfer-Learning-Accuracy}Graphs of unseen category
    recognition accuracy vs. training set size for various held-out categories.
    The rapid plateau shows that we need only a small number of examples to
    define a previously-unseen category. The accuracy difference between
    feature sets shows that the attributes are contributing significant novel information.
    Even when the attributes do not outperform material probabilities on their
    own, the combination is still superior demonstrating the rich
    discriminative information carried by the extracted material attributes.}
\end{figure*}

If perceptual material attributes are present in the material classification
network, we must be able to extract them without compromising the network's
ability to recognize materials. We compare local material recognition accuracy
with and without the auxiliary attribute loss functions to verify this. We find
that the average material category accuracy does not change when the attribute
layers are removed.  While the attribute layers are auxiliary, they are
connected to spatial pooling layers at every level and thus the  attribute
constraints affect the entire network.  If the attributes were not in fact
encoding visual material properties, constraining the network to extract them
would negatively affect the material recognition performance.

A full semantic segmentation framework is beyond the scope of this paper. We
are, however, able to use the same attribute/material CNN to produce per-pixel
material probability predictions. Results in
Fig.~\ref{fig:Material-Probability-Maps} show that we may still generate
reasonable material probability maps even from purely local information.

\section{\label{sec:Novel-Material-Category}Unseen Material Category Recognition}

One prominent application of attributes is in unseen category recognition tasks.
Examples of these tasks include one-shot~\cite{Fei-Fei2006} or zero-shot
learning~\cite{Lampert2009}. Zero-shot learning allows recognition of a unseen
category from a human-supplied list of applicable semantic attributes. Since
our attributes are non-semantic, zero-shot learning is not applicable here. We
may, however, investigate the generalization of our attributes through a form
of N-shot learning in which we use image patches extracted from a very small
number of images to learn an unseen category. While materials have been used in
the past as attributes for zero- or one-shot learning, we show that the
perceptual attributes of those materials are discriminative enough to recognize
previously-unseen materials given only a few examples.

To evaluate unseen category recognition from perceptual material attributes, we  
train a set of MAC-CNNs on modified datasets where each is missing all examples 
of a single held-out category. No examples of that category are present 
during training. The corresponding row of the category-attribute matrix is also 
removed. The same number of attributes are defined based on the remaining 
categories.

For unseen category training, we show that we require only a very
small set of examples to recognize an unseen category. We train a
simple linear binary SVM to
distinguish between the previously-seen training categories and the held-out
category based on their discovered attribute probabilities, computed on patches
extracted from each input image. We measure the effectiveness of unseen category
recognition by the fraction of final held-out category samples properly
identified as belonging to that category. As a baseline, we use the material
probability outputs from the MAC-CNN as a feature instead of attributes.

Fig.~\ref{fig:Transfer-Learning-Accuracy} shows plots of unseen category
recognition effectiveness as the number of training examples for the held-out
category varies. We can see that the accuracy plateaus quickly, indicating that
the attributes provide a compact and accurate representation for novel material
categories. The number of images we are required to extract patches from to
obtain reasonable accuracy is generally quite small (on the order of 10)
compared to full material category recognition frameworks which require
hundreds of examples. Furthermore, we include accuracy for the same predictions
based on only material probabilities instead of attribute probabilities, as
well as using a concatenation of both. Attributes alone offer better
recognition for some unseen categories. Even when they do not, the addition of
attributes still increases performance. This clearly shows that the extracted
attributes can expose discriminative information that would not
ordinarily be available.

\section{\label{sec:Conclusion}Conclusion}

We have proposed a single framework that integrates weakly-supervised attribute 
discovery with local material recognition. Our proposed CNN architecture allows 
us to discover perceptual material attributes within a local material
recognition network. To evaluate the framework, and to address issues present in 
existing  material recognition databases, we built a new material image database 
from  carefully-chosen material categories. The accuracy of unseen category 
recognition  based solely on our discovered attributes and few sample images 
shows that the  attributes form a compact representation for novel materials.

We find the parallels between our own human visual perception of materials and
the material attributes discovered in the MAC-CNN architecture particularly
interesting. Our integration of attribute and category recognition with a
single network likely has implications in other tasks such as object and scene
recognition, and we may find similar parallels there as well.

\section*{Acknowledgments}
This work was supported by the Office of Naval Research grant N00014-16-1-2158
(N00014-14-1-0316)  and N00014-17-1-2406, and the National Science Foundation
award IIS-1421094. The Titan X used for part of this research was donated by
the NVIDIA Corporation.

\bibliographystyle{ieee}
\bibliography{iccv2017}

\end{document}